\documentclass[letterpaper, 10 pt, conference]{ieeeconf}  

\IEEEoverridecommandlockouts                              

\usepackage{mathtools}
\usepackage{gensymb}
\usepackage[export]{adjustbox}
\usepackage{color}
\usepackage{graphics} 
\usepackage{epsfig} 
\usepackage{times} 
\usepackage{amsmath} 
\usepackage{tabularx}
\usepackage{amssymb}  
\usepackage{leftidx}
\usepackage{tablefootnote}
\usepackage[pagebackref]{hyperref}
\usepackage{cite}
\hypersetup{
    colorlinks=true,
    citecolor=ForestGreen,
    linkcolor=blue,
    filecolor=magenta,      
    urlcolor=cyan
}
\usepackage{booktabs}
\usepackage{multirow}
\usepackage{multicol}
\usepackage{graphicx}
\usepackage{color}
\usepackage[dvipsnames]{xcolor}
\usepackage[font=small]{caption}
\usepackage{xfrac}
\usepackage{subcaption}     

\usepackage{adjustbox}

\title{\LARGE \bf
 DiffPrompter: Differentiable Implicit Visual Prompts for Semantic-Segmentation in Adverse Conditions
}

\author{Sanket Kalwar$^{*1}$, Mihir Ungarala$^{*1}$, Shruti Jain$^{*1}$, Aaron Monis$^{1}$,   Krishna Reddy Konda$^{3}$ \\ Sourav Garg$^{2}$, K Madhava Krishna$^{1}$
\thanks{* denotes equal contribution}
\thanks{$^{1}$are with RRC, IIIT Hyderabad, India {\tt\small \{sankethkalwar, mihir.ungarala, shrutipraveenjain, aaronmonis22\}@gmail.com, mkrishna@iiit.ac.in}}%
\thanks{$^{2}$is with the Australian Institute for Machine Learning, University of Adelaide, Australia
        {\tt\small sourav.garg@adelaide.edu.au}}%
\thanks{$^{3}$is with ZF TCI, Hyderabad, India
        {\tt\small krishna.konda@zf.com}}
\thanks{$^\dagger$Code: \href{https://github.com/DiffPrompter/diff-prompter}{https://github.com/DiffPrompter/diff-prompter}}
}

\makeatletter
\let\@oldmaketitle\@maketitle
\renewcommand{\@maketitle}{\@oldmaketitle
\centering
\vspace{-2mm}

\includegraphics[width=\textwidth,height=0.4\textwidth]{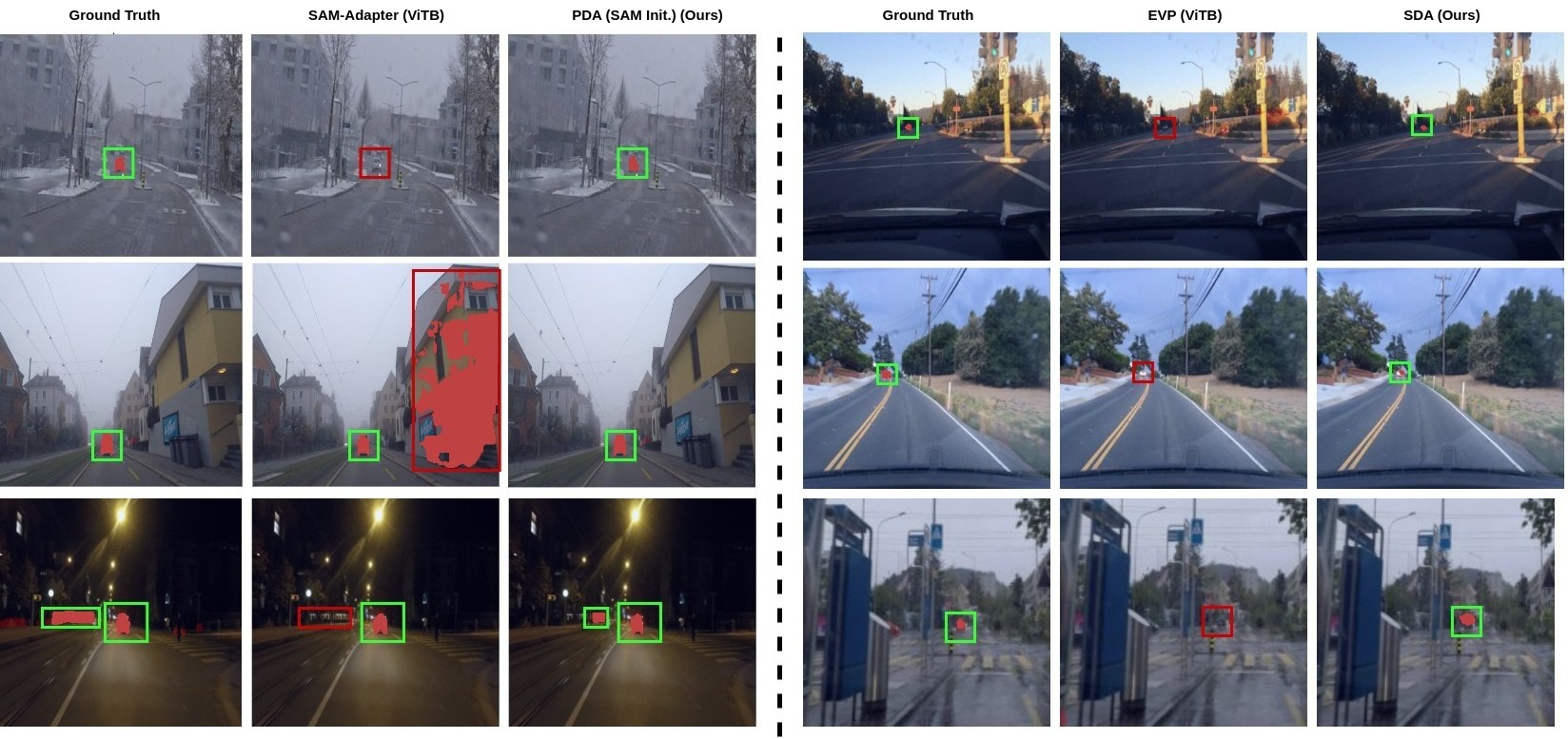}
\vspace{-6mm}
\captionof{figure}{\label{fig:qualitative_results}\small{ \textbf{Overview}: 
To tackle Semantic Segmentation in adverse weather conditions, we introduce the DiffPrompter framework (explained later in Fig.~\ref{fig:diffprompter}), through which we create a \textit{Serial} Differentiable Adaptor (SDA) and \textit{Parallel} Differentiable Adaptor (PDA). Both these adaptors achieve superior results as compared to the current state-of-the-art (SOTA) methods: SAM-Adapter~\cite{samAdapt} (left) and EVP \cite{evp} (right), where columns in each case respectively represent the ground-truth segmentation mask, the baseline method (SAM-Adapter/EVP), and our proposed method (PDA/SDA). In this representation, the model's masked output is shown in red, with green bounding boxes highlighting correct predictions and red bounding boxes indicating missed segmentation outputs (false negatives) or incorrect segmentation outputs (false positives). It is evident that our proposed methods, PDA and SDA, demonstrate superior qualitative performance compared to the SOTA methods, EVP and SAM-Adapter}.}

\label{fig:qualitative}

}

\begin{document}

\maketitle
\thispagestyle{empty}
\pagestyle{empty}



\begin{abstract}
Semantic segmentation in adverse weather scenarios is a critical task for autonomous driving systems. While foundation models have shown promise, the need for specialized adaptors becomes evident for handling more challenging scenarios. We introduce \textit{DiffPrompter}, a novel differentiable visual and latent prompting mechanism aimed at expanding the learning capabilities of existing adaptors in foundation models. Our proposed \textit{$\nabla$HFC} (High Frequency Components) based image processing block excels particularly in adverse weather conditions, where conventional methods often fall short. Furthermore, we investigate the advantages of \textit{jointly training visual and latent prompts}, demonstrating that this combined approach significantly enhances performance in out-of-distribution scenarios. Our differentiable visual prompts leverage \textit{parallel} and \textit{series} architectures to generate prompts, effectively improving object segmentation tasks in adverse conditions. Through a comprehensive series of experiments and evaluations, we provide empirical evidence to support the efficacy of our approach. Project page: \href{https://diffprompter.github.io}{diffprompter.github.io} 
\end{abstract}

%
\section{INTRODUCTION}
Scene understanding or semantic segmentation for Autonomous Vehicles (AV) is an important part of their perception stack through which they can recognize the map elements, open space and other vehicles and obstacles. Most deep-learning approaches for semantic segmentation work well in ideal environmental conditions \cite{cordts2016cityscapes} but fail under adverse conditions such as low light, fog, snow, rain and glare \cite{yu2020bdd100k, sakaridis2021acdc, Zendel_2018_ECCV, Zendel_2022_CVPR ,sakaridis2019guided}. 

In this paper, we tackle semantic segmentation in adverse conditions by leveraging large foundation models and fine-tuning them using \textit{differentiable} visual prompts. While these large models perform well in a zero-shot manner in a variety of domains, they still require fine-tuning to generalize on adverse environmental conditions. Some approaches focus on fully fine-tuning a pre-trained model on the downstream task \cite{ ding2023parameter}. However, this strategy requires one to store and deploy a separate copy of the backbone parameters for every single task. This is computationally expensive and infeasible in most real-time scenarios. Other strategies \cite{hu2023efficiently} include fine-tuning only the last few layers or the neck of the model but these methods underperform as compared to complete fine-tuning of the model. 

Contrary to the above, we take inspiration from~\cite{adapterRef1,hu2023llmadapters,adapterRef2, Sandler_2022_CVPR, fu2023effectiveness} which introduced additional residual blocks (or Adaptors) for the network backbone. In these architectures, the backbone is kept frozen and only the Adaptors are trained,  making it a lightweight way to finetune large models while performing on par with other finetuning techniques. 
To further boost the performance, \cite{evp,samAdapt} introduced visual prompts as inputs to the Adaptors. These methods work well because they introduce task-specific information through Adaptors while retaining the knowledge learned by the large models. However, these methods fail to generalize well in out-of-distribution settings because the visual prompts fail to adapt to new domains and new data distributions.


To address this issue, we use \textit{differentiable visual prompts} which are also learned during training. The prompts focus on enhancing the image using differentiable parameters as demonstrated by~\cite{gdip}. A similar idea has also been explored in~\cite{evpUniversal} where they introduced a differentiable FFT block as a visual prompt to the Adaptor. With our differentiable image processing block, we provide an array of image pre-processing techniques that can be combined with each other to produce a more meaningful image prompt for any domain. 

The contributions of this paper are listed below:
\begin{enumerate}
    \item a \textit{differentiable} visual prompting mechanism to extend the generalizing abilities in out-of-distribution settings of existing adaptors for foundation models (see Table \ref{tab:table_bdd100k_qualitative}); 
    \item a $\nabla HFC$ (High Frequency Components) retaining differentiable image processing block, which is particularly robust to adverse conditions (see Sec.~\ref{sec:abalation} and Table~\ref{tab:hfc_abalation});
    \item \textit{joint learning} of visual prompts and latent prompts, which outperforms their individually-learnt versions (see Sec.~\ref{sec:abalation} and Table~\ref{tab:table_dvp_abalation}); and
    \item \textit{parallel} and \textit{serial} versions of our architecture to generate visual and latent prompts that improve the object-segmentation task in adverse conditions (see Fig.~\ref{fig:qualitative_results}, Table \ref{tab:table_bdd100k_qualitative_bdd_only},~\ref{tab:table_bdd100k_qualitative},~\ref{tab:table_cod_result} and ~\ref{tab:table_cod_camo_cham_result}).
\end{enumerate}


\section{RELATED WORK}
\label{sec:related_work}

\noindent \textbf{Foundation Segmentation Models:}
Semantic segmentation task is often tackled using methods like Fully Convolutional Networks (FCNs) \cite{FCN}, encoder-decoder structures \cite{unet, segnet}, atrous convolutions \cite{deeplab, Chen_2018_ECCV}, pyramid modules \cite{pspnet, statisticaltexture} and attention modules \cite{DANnet, ccnet, nonlocalnet}.
More recently, transformer-based methods have been shown to be effective in semantic segmentation \cite{semask, segmenter, SETR}. 
SETR \cite{SETR} was one of the first to remodel image segmentation as a transformer-based sequence-to-sequence prediction task.
Segformer \cite{segformer} uses a novel hierarchically structured transformer encoder and a decoder that aggregates this information allowing for high performance even on zero-shot distributions.
The Segment Anything Model (SAM) \cite{kirillov2023segment} introduces a foundational model in the form of a ViT-based model trained upon a large visual corpus. It has been designed and trained to be transferred zero-shot to new image distributions and tasks. It produces high-quality image masks from input prompts like points or boxes and can be used to generate masks for all objects in an image. 

\noindent \textbf{Visual Prompt tuning:}
Prompting was initially introduced in NLP \cite{brown2020language, liu2023pre}. \cite{brown2020language} demonstrates that GPT-3 shows strong generalization to downstream tasks even in the few-shot or zero-shot settings with manually chosen prompts. Recent works have introduced prompting to vision tasks \cite{vpt, evp, evpUniversal,Sandler_2022_CVPR, samAdapt}. \cite{Sandler_2022_CVPR} introduces a set of learnable embedding vectors at each layer of a transformer. Visual Prompt Tuning (VPT) \cite{vpt} modifies the input to the Vision Transformer by introducing a small amount of task-specific learnable parameters into the input space by simply prepending into the input sequence and are learned together using a linear head during fine-tuning.
Explicit Visual Prompting \cite{evp} introduces tunable parameters that extract explicit visual information from frozen patch embeddings and the high-frequency components of input images to create explicit prompts. The explicit prompt created from these features is used as an Adaptor for each layer of the ViT.
\renewcommand{\thefigure}{2}
\begin{figure}
    \centering
    \includegraphics[width=0.5\textwidth]{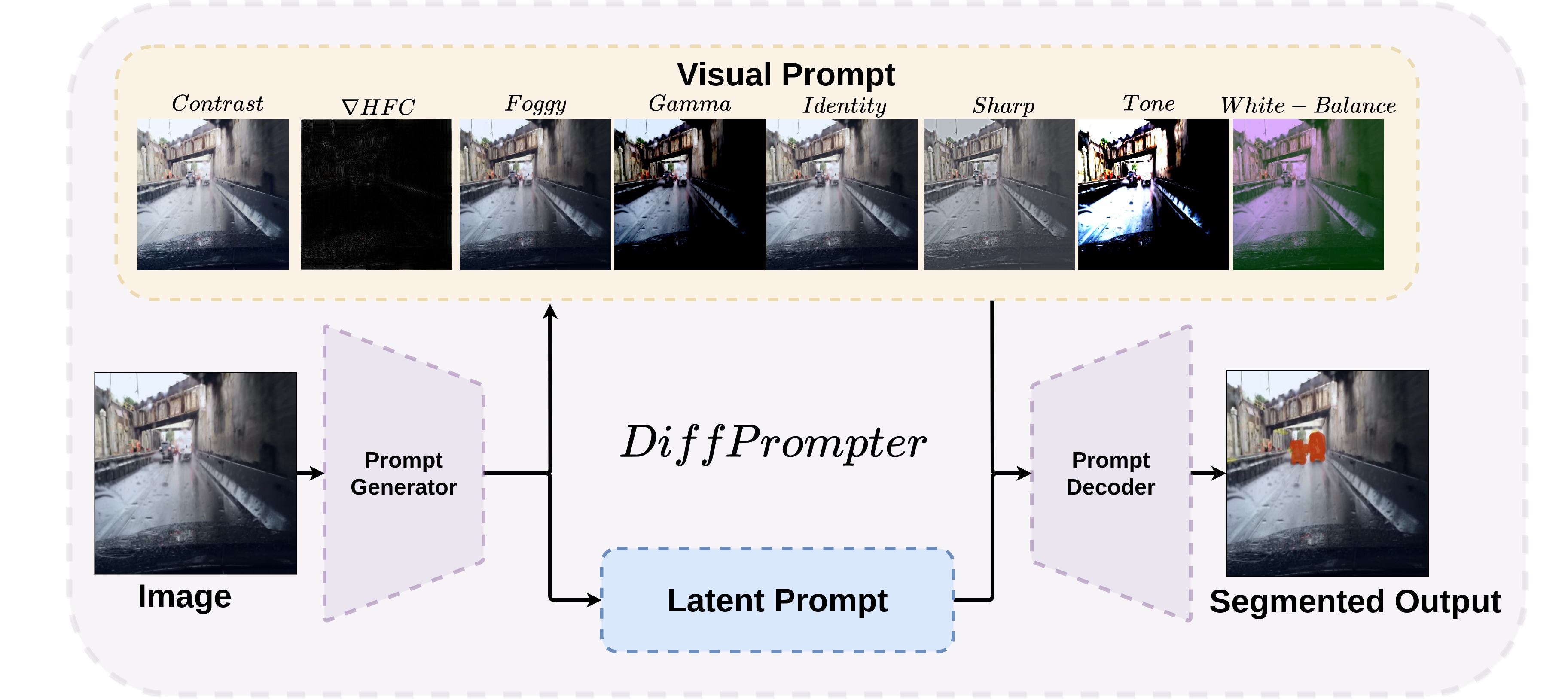}
    \caption{We propose \textbf{DiffPrompter}, a semantic segmentation method that utilizes the visual and latent prompts generated by the prompt generator. These prompts are then used by the prompt decoder to generate a semantic mask for segmenting objects, especially in adverse conditions and low-level segmentation tasks.}
    \label{fig:diffprompter}
\end{figure}

\noindent \textbf{Differentiable Image enhancement:}
Image enhancement is widely used to provide better features for vision tasks. 
Exposure \cite{hu2018exposure} learns to apply a sequence of enhancement operations by using deep Reinforcement Learning. IA-Yolo \cite{liu2022imageadaptive} uses a CNN to predict differentiable image processing parameters trained in conjunction with Yolov3 \cite{redmon2018yolov3} for object detection in an end-to-end fashion. 
GDIP \cite{gdip} proposes a domain-agnostic novel gating mechanism for learning parameters of multiple image pre-processing techniques that operate concurrently. This mechanism can be plugged into existing object detection networks like YOLOv3 \cite{redmon2018yolov3} and trained end-to-end on adverse condition images. \cite{evpUniversal, evp} extracts the high-frequency components of an image to create the explicit visual prompts.

Unlike the methods mentioned above, our method uses differentiable visual prompting for the adaptors on foundation models. For the visual prompt, we use our $\nabla HFC$ block along with other image-processing techniques mentioned in \cite{gdip}. In contrast to previous methods, we jointly learn visual and latent prompts in a parallel and serial manner. 

\renewcommand{\thefigure}{3}
\begin{figure*}[htb!]
\centering
\includegraphics[width=\textwidth,height=.3\textwidth]{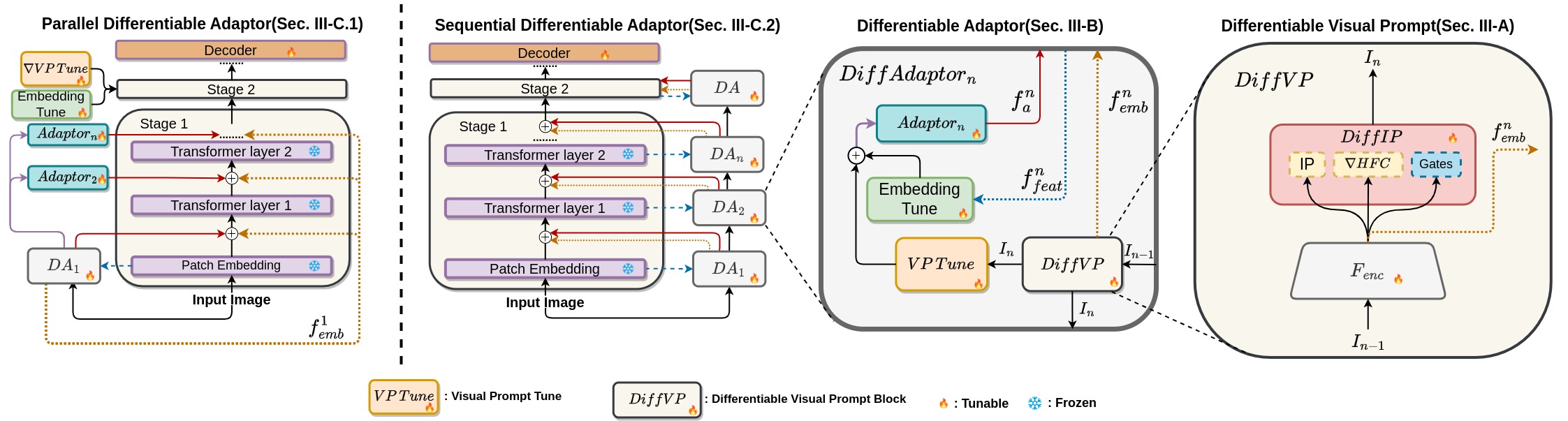}
\caption{Framework overview, First, the input image is sent to a Parallel Differentiable Adaptor ($PDA$, Sec.\ref{sec:pda}) and Sequential Differentiable Adaptor 
 ($SDA$, Sec.\ref{sec:sda}) segmentation model through various stages of transformer encoder and decoder layers which in turn generates semantic segmentation mask as output, details in Sec.\ref{sec:psm}, $DiffAdaptor$ is used as a building block in both $PDA$ and $SDA$ which contains $DiffVP$ block that generates differentiable visual prompt details of both are discussed in Sec.\ref{sec:da} and Sec.\ref{sec:dvp}.}
\label{fig: dvp_proposed_method}
\end{figure*}
\section{PROPOSED METHOD}


We propose a Differentiable Visual Prompt block, referred to as $DiffVP$, as shown in Fig. \ref{fig: dvp_proposed_method}. This block learns visual prompts using $DiffIP$ and latent prompts through a shallow vision encoder. The Differentiable Adaptor, denoted as $DiffAdaptor$, employs $VPTune$ to learn local information of the visual prompt. Local information from the transformer layer of the encoder is provided to the $Embedding\,\,Tune$ layer. The output of the $Embedding\,\,Tune$ and $VPTune$ layers is combined and fed into the adaptor layer, which outputs local features that when added to the transformer layers, learn local invariant features. To introduce global invariance in the features, the latent embedding generated by the shallow vision encoder is added to the transformer output features. This combination of local and global invariance aids the Parallel Differentiable Adaptor (PDA) and Sequential Differentiable Adaptor (SDA) in improving semantic segmentation tasks.


\label{sec:proposed_method}

\subsection{\textbf{Differentiable Visual Prompt ($DiffVP$) Block:}}
\label{sec:dvp}
We utilize a shallow vision encoder, denoted as $F_{enc}$, in combination with the differential image processing block $DiffIP$, which incorporates our proposed $\nabla$HFC image processing operation. This combination is used to generate the visual prompt output $I_{n}$ along with the output latent embedding $f_{emb}^{n}$ as follows:
\begin{equation}
\begin{aligned}
    f_{emb}^{n} = \textit{F}_{enc}(\textit{I}_{n-1})
\end{aligned}
\label{eq:f_emb}
\end{equation}

\begin{equation}
\begin{aligned}
    \textit{I}_{n} = DiffIP(f_{emb}^{n})
\end{aligned}
\label{eq:diffip}
\end{equation}

The $n$ indicates the total number of transformer layers in the ViT-B encoder discussed in Section \ref{sec:psm}. In the section below, we will delve into the building blocks of $DiffVP$, which include the shallow vision encoder, the $\nabla$HFC image processing block, and $DiffIP$.

\subsubsection{\textbf{Shallow Vision Encoder}}
This lightweight vision encoder, denoted as $F_{enc}$, takes an image $I \in R^{H\times W \times 3}$ as input and learns a latent embedding $f_{emb}$ that encapsulates the global context of the image. This embedding is valuable for predicting the parameters of the image processing blocks in $DiffIP$. The encoder comprises five consecutive 2D convolution layers with a kernel size of 3 and a stride of 1, along with the \textit{ReLU} non-linearity. Finally, adaptive average pooling is applied to reduce the spatial dimensions, followed by a fully connected layer that produces the latent embedding $f_{emb}$. This process can be expressed using Eq.\ref{eq:f_emb}.

\subsubsection{\textbf{$\nabla$HFC Image Processing Block}}
We propose a method to learn high-frequency components directly by parameterizing the HFC component of the image as mentioned in the work \cite{evp}, where low-frequency components of the image are removed and high-frequency components are kept by masking the $z=fft(I)$ by the mask $\textbf{M}_{hfc} \in \{0,1\}^{H\times W}$ 
 which retains high-frequency components of $z$ and that this retained component is inverse Fourier transformed to get the high-frequency component image as $I_{hfc}=ifft(z \textbf{M}_{hfc})$. In \cite{evp}  $\tau$ rate parameter is tuned manually for the given dataset. However, this parameter can be made learnable by predicting it directly using shallow vision encoder output $f_{emb}$ along with $F_{hfc}$ fully-connected layer and sigmoid non-linearity $\sigma$ as shown in the Eq.\eqref{eq:tau} denoted by $\Tilde{\tau}$.
\begin{equation}
\begin{aligned}
    \Tilde{\tau} = max(0.1,min(\sigma(\textit{F}_{hfc}(f_{emb})),0.8))\\
\end{aligned}
\label{eq:tau}
\end{equation}
\begin{equation}
\begin{aligned}
    \Tilde{\textbf{M}}_{\nabla hfc}^{i,j} = 
    \begin{cases}
      0, & \text{if}\ \frac{HW-4|(i-\frac{H}{2})(j-\frac{W}{2})|}{HW} < \Tilde{\tau}\\
      1, & \text{otherwise}
    \end{cases}
\label{eq:masks}
\end{aligned}
\end{equation}
\begin{equation}
\begin{aligned}
    \Tilde{I}_{\nabla hfc} = ifft(z\Tilde{\textbf{M}}_{\nabla hfc})
\label{eq:dfft_image}
\end{aligned}
\end{equation}

Mask $\Tilde{\textbf{M}}_{\nabla hfc}^{i,j}$ can be further generated using $\Tilde{\tau}$ learned during training as shown in the Eq. \eqref{eq:masks}, which enables learning high-frequency component image $\Tilde{I}_{\nabla hfc}$ directly in end-to-end fashion by the process mentioned in Eq. \eqref{eq:dfft_image}.  

\subsubsection{\textbf{Differentiable Image Processing Block ($DiffIP$)}}
We propose a new image processing operation, denoted as $\nabla$HFC, for extracting high-frequency components from an image. This operation is introduced in addition to the image processing techniques utilized in GDIP model~\cite{gdip}, which includes tone adjustment, contrast enhancement, sharpening, defogging, gamma correction, white balance, and identity operation. When these image processing operations are combined, they form a new block known as $DiffIP$. The $DiffIP$ block consists of image-processing operations represented as $ip$ and their corresponding gates denoted as $w_{ip}$. The parameters for these operations are predicted using $f_{emb}$, which is obtained from a shallow vision encoder. The $DiffIP$ block can be succinctly summarized by combining Eq.\ref{eq:f_emb} and Eq.\ref{eq:dip}, where $f_{ip}$ represents the output image resulting from the image processing operation $F_{ip}$, and $N$ denotes a min-max normalization operation.
\begin{equation}
\begin{aligned}
    I_{n} = N(\sum_{ip} w_{ip} N(F_{ip}(f_{emb}^{n})))
\label{eq:dip}
\end{aligned}
\end{equation}

\subsection{\textbf{Differentiable Adaptor ($DiffAdaptor$):}}
\label{sec:da}
The Differentiable Adaptor takes image $I_{n-1}$ as input, which is initially passed through the $DiffVP$ block. This block outputs a differentiable visual prompt $I_{n}$ and $f_{emb}^n$, and the $I_{n}$ is utilized by the $VPTune$ layer. The $VPTune$ layer performs patch-wise convolution to generate patch embeddings $f_{VPTune}$. The embedding layer takes the features $f_{feat}^{n}$ from the corresponding $n^{th}$ layer of the ViT encoder and generates the output embedding $f_{embedding\_tune}$. This output embedding is then added to the patch embeddings generated by the $VPTune$ layer and is finally fed into the adaptor layer.\\
The adaptor block is composed of two components: a shared MLP denoted as $MLP_{shared}$ and unshared MLPs represented as $MLP_{n}$ for each of the transformer layers at all stages of the encoder. This composition can be succinctly expressed using Equations \ref{eq:da_x} and \ref{eq:da_f_a}.
\begin{equation}
\begin{aligned}
   x = f_{VPTune}+f_{embedding\_tune}
\label{eq:da_x}
\end{aligned}
\end{equation}
\begin{equation}
\begin{aligned}
    f_{a}^n = MLP_{shared}(GELU(MLP_{n}(x)))
\label{eq:da_f_a}
\end{aligned}
\end{equation}

\subsection{\textbf{Parallel-Sequential Differentiable Adaptor Models:}}
\label{sec:psm}
PDA and SDA comprise a ViT-B encoder and a progressive upsampling ConvNet layer as its decoder, as illustrated in Fig. \ref{fig: dvp_proposed_method}. In the ViT encoder, all transformer layers remain frozen at every stage, while the decoder, Adaptor layers, and the $DiffAdaptor$ units are fine-tuned during training. In the section below, we will delve into the details of PDA and SDA.
\subsubsection{\textbf{Parallel Differentiable Adaptor Model (PDA)}}
\label{sec:pda}
The PDA encoder follows a standard forward pass procedure. The input image is processed through various transformer layers and ultimately reaches the decoder layer. Additionally, at each skip connection, features computed from the Adaptor, along with the $f_{emb}$ features, are incorporated. In the initial stage of the ViT-B encoder, only one $DiffAdaptor$ is employed. Its output is computed by $DiffVP$, which takes the input image and generates the visual prompts, along with the latent prompt $f_{emb}$. Subsequently, this output is fed into $VPTune$, which utilizes the visual prompts, and $Embedding\,\,Tune$, which takes patch embeddings as input. The sum of the outputs from $VPTune$ and $Embedding\,\,Tune$ is then provided to all the Adaptor layers, the details of which are outlined in Sec. \ref{sec:da}.
\\
In PDA, the computations for $DiffVP$ are performed only once and are concurrently utilized by all the Adaptor layers associated with that particular stage.

\subsubsection{\textbf{Sequential Differentiable Adaptor Model (SDA)}}
\label{sec:sda}
In contrast to the PDA model, the SDA incorporates $DiffAdaptor$ blocks at all stages of the encoder. The outputs of the $DiffAdaptor$, namely the latent prompt $f_{emb}$ and the Adaptor output $f_{a}^{n}$, are added to the skip-connection (plus sign between the transformer layer in the Fig.\ref{fig: dvp_proposed_method}) of the ViT-B encoder. Additionally, the $I_{n}$ image is generated as an output by the $DiffAdaptor$ and subsequently serves as an input to the next $DiffAdaptor$ block in a sequential manner.

\section{EXPERIMENTAL SETUP}
\label{sec:exp_setup}
\subsection{Datasets}

For the vehicle foreground segmentation task, we use \textbf{BDD100K} dataset \cite{yu2020bdd100k} (approximately 7,000 train and 1,000 test images), \textbf{ACDC} test dataset \cite{sakaridis2021acdc} (406 images), \textbf{Wild-Dash} test dataset \cite{Zendel_2018_ECCV, Zendel_2022_CVPR} (70 images) and \textbf{Dark-Zurich} test dataset \cite{sakaridis2019guided} (50 images). These datasets specialize in semantic segmentation in adverse weather conditions. BDD100K specializes in scenarios with rain, snow, clear, overcast, foggy and cloudy conditions. ACDC dataset has conditions with fog, nighttime, rain and snow. Wild-Dash comprises real-world driving scenarios in challenging conditions and Dark-Zurich consists of images in nighttime, twilight and daytime.
Focusing mainly on vehicle segmentation, we combine the vehicle classes in the above-mentioned datasets, as a single foreground class, while the rest are considered as a background class. 
In the BDD100K dataset, out of the 19 classes, we use 6 classes for foreground - car, truck, bus, train, motorcycle, and bicycle.
Similarly, in ACDC, Wild-Dash and Dark-Zurich datasets, from the 34 labels in the annotated data, we use 8 classes for foreground i.e. car, truck, bus, caravan, trailer, train, motorcycle and bicycle.

For the low-level segmentation task we use \textbf{COD10K} \cite{fan2020camouflaged} and \textbf{CAMO} \cite{ltnghia-CVIU2019} dataset. COD10K is the largest dataset for camouflaged object detection, which contains 3,040 training and 2,026 testing samples. CAMO provides diverse images with naturally camouflaged objects and artificially camouflaged objects, which contains 1,000 training and 250 test samples. 
Following the training protocol from EVP \cite{evp} and SAM-Adapter \cite{samAdapt}, both COD10k and CAMO datasets are combined, but while testing, they are evaluated independently on their respective test sets.
\begin{table}[!htb]
\caption{\label{tab:table_bdd100k_qualitative_bdd_only} Comparison between EVP-Net and SAM-Adapter with our proposed  PDA/SDA trained on BDD100k trainset and tested on BDD100k testset. The colors denoted in \textcolor{green}{green} and \textcolor{red}{red} correspond to a relative increase or decrease in the performance of PDA and SDA, respectively, in comparison to the baseline methods.}
\begin{adjustbox}{width=0.48\textwidth}
\begin{tabular}{|l|llll|}
\hline
\multicolumn{1}{|c|}{\multirow{2}{*}{Method}} & \multicolumn{4}{c|}{BDD100K}                                                                 \\
\multicolumn{1}{|c|}{}                        & \multicolumn{1}{l|}{$S_{\alpha} \uparrow$}     & \multicolumn{1}{l|}{$E_{\phi}\uparrow$}     & \multicolumn{1}{l|}{$F_{\beta}^w \uparrow$}    & $MAE\downarrow$         
\\ \hline
EVP-Net                                & \multicolumn{1}{l|}{.8466} & \multicolumn{1}{l|}{.9225} & \multicolumn{1}{l|}{.7521} & .03033
\\ \hline
PDA(Ours)                     & \multicolumn{1}{l|}{.8513 \textcolor{green}{(+0.56)}}    & \multicolumn{1}{l|}{.9274 \textcolor{green}{(+0.53)}}    & \multicolumn{1}{l|}{.7590 \textcolor{green}{(+0.92)}}    & \textbf{.0294} \textcolor{green}{(-3.1)}       
\\ \hline
SDA(Ours)                     & \multicolumn{1}{l|}{\textbf{.8523} \textcolor{green}{(+0.67)}}    & \multicolumn{1}{l|}{\textbf{.9294 } \textcolor{green}{(+0.75)}}    & \multicolumn{1}{l|}{\textbf{.7624 } \textcolor{green}{(+1.37)}}    & .0296 \textcolor{green}{(-2.4)}  
\\ \hline\hline
SAM-Adapter                            & \multicolumn{1}{l|}{.9222} & \multicolumn{1}{l|}{.9561} & \multicolumn{1}{l|}{.8571} & .0155  
\\ \hline
PDA(Ours)                        & \multicolumn{1}{l|}{.9251 \textcolor{green}{(+0.31)}}    & \multicolumn{1}{l|}{.9592 \textcolor{green}{(+0.32)}}    & \multicolumn{1}{l|}{\textbf{.8656} \textcolor{green}{(+0.99)}}    & .0144 \textcolor{green}{(-7.63)}
\\  \hline
SDA(Ours)                        & \multicolumn{1}{l|}{\textbf{.9279} \textcolor{green}{(+0.62)}}    & \multicolumn{1}{l|}{\textbf{.9653} \textcolor{green}{(+0.96)}}    & \multicolumn{1}{l|}{.8495 \textcolor{red}{(-0.89)}}    & \textbf{.0100 }\textcolor{green}{(-55.10)}    
\\ \hline
\end{tabular}
\end{adjustbox}
\end{table}

\begin{table*}[htb]
\caption{\label{tab:table_bdd100k_qualitative} Comparison of EVP-Net and SAM-Adapter with our proposed PDA/SDA on ACDC, Wild-Dash and Dark-Zurich dataset. Note that the training set only comprises BDD100K and is evaluated on the ACDC, Wild-Dash and Dark-Zurich for testing the generalization ability of the models.}
\scalebox{0.68}{
\begin{tabular}{|l|llll||llll||llll|}
\hline
\multicolumn{1}{|c|}{\multirow{2}{*}{Method}} & \multicolumn{4}{c||}{ACDC}                                                                        & \multicolumn{4}{c||}{Wild-Dash}                                                            & \multicolumn{4}{c|}{Dark-Zurich}                                                                 
\\
\multicolumn{1}{|c|}{}                        & \multicolumn{1}{l|}{$S_{\alpha} \uparrow$}     & \multicolumn{1}{l|}{$E_{\phi}\uparrow$}     & \multicolumn{1}{l|}{$F_{\beta}^w \uparrow$}    & $MAE\downarrow$     & \multicolumn{1}{l|}{$S_{\alpha} \uparrow$}     & \multicolumn{1}{l|}{$E_{\phi}\uparrow$}     & \multicolumn{1}{l|}{$F_{\beta}^w \uparrow$}    & $MAE\downarrow$    & \multicolumn{1}{l|}{$S_{\alpha} \uparrow$}     & \multicolumn{1}{l|}{$E_{\phi}\uparrow$}     & \multicolumn{1}{l|}{$F_{\beta}^w \uparrow$}    & $MAE\downarrow$      \\ \hline

EVP-Net & \multicolumn{1}{l|}{.6859} & \multicolumn{1}{l|}{.7710} & \multicolumn{1}{l|}{.4298} & .0249 & \multicolumn{1}{l|}{.7409} & \multicolumn{1}{l|}{.8218} & \multicolumn{1}{l|}{.3741} & .0297 & \multicolumn{1}{l|}{.6202} & \multicolumn{1}{l|}{.6037} & \multicolumn{1}{l|}{.1766} & .0194 \\ \hline
PDA(Ours) & \multicolumn{1}{l|}{.6941 \textcolor{green}{(+0.1)}}     & \multicolumn{1}{l|}{.7755 \textcolor{green}{(+0.58)}}    & \multicolumn{1}{l|}{.4406 \textcolor{green}{(+2.51)}}    & .0245 \textcolor{green}{(-1.6)}    & \multicolumn{1}{l|}{.7289 \textcolor{red}{(-1.64)}}    & \multicolumn{1}{l|}{.7996 \textcolor{red}{(-2.78)}}    & \multicolumn{1}{l|}{.3627 \textcolor{red}{(-3.14)}}    & .0317  \textcolor{red}{(+6.7)}    & \multicolumn{1}{l|}{.6245 \textcolor{green}{(+0.69)}}    & \multicolumn{1}{l|}{.6216 \textcolor{green}{(+2.97)}}    & \multicolumn{1}{l|}{.1903 \textcolor{green}{(+7.76)}}    & .0169 \textcolor{green}{(-14.7)}   \\ \hline
SDA(Ours)  & \multicolumn{1}{l|}{\textbf{.6948}  \textcolor{green}{(+1.3)}}    & \multicolumn{1}{l|}{\textbf{.7886} \textcolor{green}{(+2.28)}}    & \multicolumn{1}{l|}{\textbf{.4542} \textcolor{green}{(+5.68)}}    & \textbf{.0233} \textcolor{green}{(-6.8)}     & \multicolumn{1}{l|}{\textbf{.7480} \textcolor{green}{(+0.95)}}    & \multicolumn{1}{l|}{\textbf{.8273} \textcolor{green}{(+0.67)}}    & \multicolumn{1}{l|}{\textbf{.3976} \textcolor{green}{(+6.28)}}    & \textbf{.0282} \textcolor{green}{(-5.3)}    & \multicolumn{1}{l|}{\textbf{.6294} \textcolor{green}{(+1.48)}}    & \multicolumn{1}{l|}{\textbf{.6573} \textcolor{green}{(+8.88)}}    & \multicolumn{1}{l|}{\textbf{.1938} \textcolor{green}{(+9.74)}}    & \textbf{.0148} \textcolor{green}{(-31.1)}   \\ \hline\hline
SAM-Adapter   & \multicolumn{1}{l|}{.8366} & \multicolumn{1}{l|}{.8871} & \multicolumn{1}{l|}{\textbf{.6756}} & .0107 & \multicolumn{1}{l|}{.8873} & \multicolumn{1}{l|}{.9207} & \multicolumn{1}{l|}{.6147} & \textbf{.0093} & \multicolumn{1}{l|}{.7033} & \multicolumn{1}{l|}{.7349} & \multicolumn{1}{l|}{\textbf{.3125}} & .0132 \\ \hline
PDA(Ours)     & \multicolumn{1}{l|}{.8448 \textcolor{green}{(+0.98)}}    & \multicolumn{1}{l|}{.8939 \textcolor{green}{(+0.77)}}    & \multicolumn{1}{l|}{.6751 \textcolor{red}{(-0.07)}}    & .0106 \textcolor{green}{(-0.94)}     & \multicolumn{1}{l|}{\textbf{.8944} \textcolor{green}{(+0.80)}}    & \multicolumn{1}{l|}{.9220 \textcolor{green}{(+0.14)}}    & \multicolumn{1}{l|}{.6346 \textcolor{green}{(+3.24)}}    & .0094  \textcolor{red}{(+0.1)}  & \multicolumn{1}{l|}{\textbf{.7100} \textcolor{green}{(+0.95)}}    & \multicolumn{1}{l|}{.7385 \textcolor{green}{(+0.49)}}    & \multicolumn{1}{l|}{.3036 \textcolor{red}{(-2.93)}}    & .0143  \textcolor{red}{(+8.34)}  \\ 
\hline
SDA(Ours)    & \multicolumn{1}{l|}{\textbf{.8460} \textcolor{green}{(+1.12)}}    & \multicolumn{1}{l|}{\textbf{.9001} \textcolor{green}{(+1.47)}}    & \multicolumn{1}{l|}{.6690 \textcolor{red}{(-0.98)}}    & \textbf{.0100}  \textcolor{green}{(-7)}     & \multicolumn{1}{l|}{.8820 \textcolor{red}{(-0.60)}}    & \multicolumn{1}{l|}{\textbf{.9310} \textcolor{green}{(+1.11)}}    & \multicolumn{1}{l|}{\textbf{.6658} \textcolor{green}{(+8.31)}}    & .0097 \textcolor{red}{(+4.3)}    & \multicolumn{1}{l|}{.6718 \textcolor{red}{(-4.69)}}    & \multicolumn{1}{l|}{\textbf{.7412} \textcolor{green}{(+0.85)}}    & \multicolumn{1}{l|}{.2627 \textcolor{red}{(-18.95)}}    & \textbf{.0087} \textcolor{green}{(-51)}   \\ \hline
\end{tabular}
}

\end{table*}

\subsection{Training Setup}
PDA and SDA models are trained on input images with a size of $352\times352\times3$ and a batch size of 4. Additionally, we employ the $AdamW$ optimizer with a $lr$ of 2e-4, a multi-step LR scheduler with a gamma parameter of 0.1, and a milestone parameter set to 1. We train PDA and SDA for 50 epochs on the BDD100K dataset and subsequently evaluate their performance on BDD100K, ACDC, Wild-Dash, and Dark-Zurich. For the CAMO and COD10K datasets, both models are trained for 20 epochs. For both setups, we use Binary Cross Entropy Loss during backpropagation.
\par PDA(SAM Init) undergoes training in a similar fashion as mentioned above, with the difference that the image dimensions are set to $1024 \times 1024 \times 3$, with a batch size of 3. Similarly, for SDA(SAM Init), the training happens in a similar fashion except that the batch size is 1. Our proposed method PDA and SDA, are constructed on the foundation of the ViT-B architecture, selected for its suitability in real-time autonomous driving applications. In pursuit of fairness and transparency, we have chosen to utilize ViT-B as the backbone for both the SAM-Adapter and EVP models, ensuring an impartial comparison with our own PDA and SDA innovations. 
\renewcommand{\thefigure}{4}
\begin{figure*}[htb!]
\includegraphics[width=\textwidth,height=.3\textwidth]{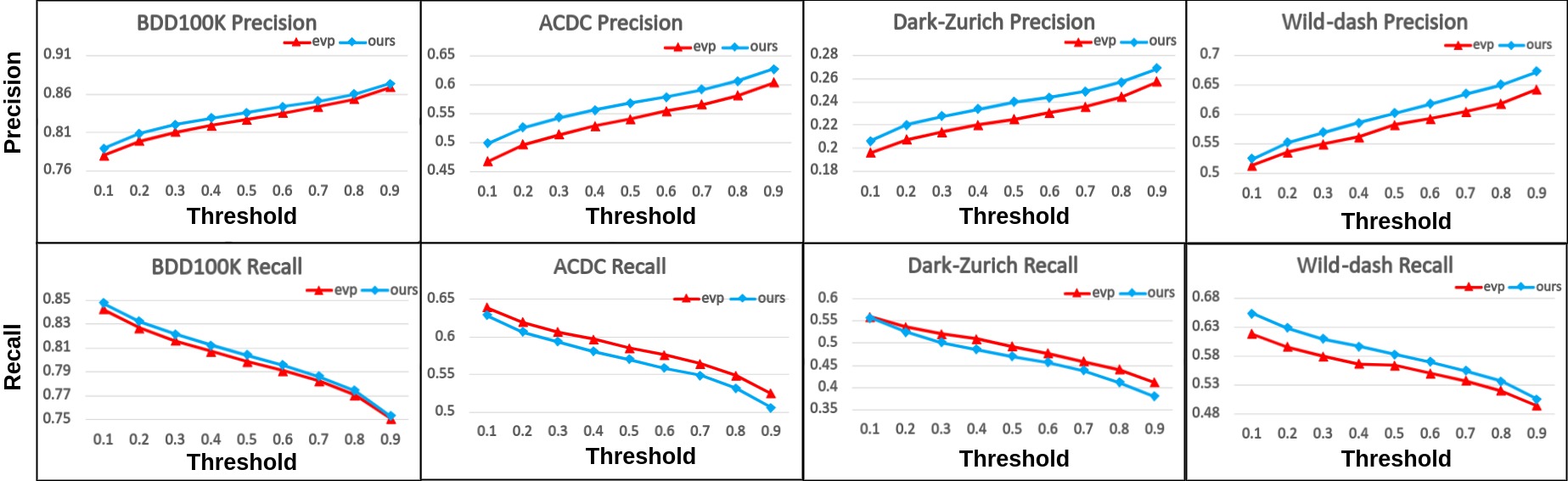}
\centering
\caption{\textbf{Precision \& Recall} graph, where row 1 corresponds to Precision and row 2 corresponds to Recall, also columns 1-4 belong to BDD100K, ACDC, Dark-Zurich and Wild-Dash datasets respectively. The red colour belongs to the EVP method while the blue colour belongs to SDA(ours).}
\label{fig:stat_plot}
\end{figure*}
\section{RESULTS AND ANALYSES}
\label{sec:results_analyses}


\subsection{Qualitative Analysis}
We compare the results of PDA and SDA with the foundation model SAM-Adapter and EVP as shown in Fig.\ref{fig:qualitative_results} for semantic segmentation on real-world data. On the left side of the vertical dashed line, SAM-Adapter has missed many instances (row 1 and row 3) or classified incorrectly (row 2) as compared to our proposed method which does better on objects far away as well as on adverse conditions images like snow, fog and night scenarios. Likewise, we see similar trends to the right side of the vertical dashed line where SDA does better qualitatively as compared to EVP on the far away objects in adverse conditions like dusk, cloudy and rainy scenarios. We quantify the improvements extended by PDA and SDA in the next section.

\subsection{Quantitative Analysis}
We compare our proposed method with the SOTA method EVP and SAM-Adapter on the BDD100K, ACDC, Wild-Dash and Dark-Zurich test dataset as shown in Table \ref{tab:table_bdd100k_qualitative_bdd_only} and Table \ref{tab:table_bdd100k_qualitative}. These results are discussed in the subsequent Sec. \ref{a} and Sec. \ref{b}. The main focus of this study is to show that our method is able to perform well on \textit{high-level} semantic segmentation for autonomous driving settings. Additionally, in Table \ref{tab:table_cod_result}, we show that our proposed architecture retains performance even for the \textit{low-level} semantic segmentation task. In Table~\ref{tab:table_cod_camo_cham_result}, we compare our proposed method PDA and SDA with the SOTA method EVP on COD10K and CAMO datasets, as discussed in Sec.~\ref{c}. We use the following evaluation metrics: S-measure $S_{\alpha}$ \cite{fan2017structure} provides similarity score for objects, background and regions structure with respect to the ground truth mask; mean E-measure $E_{\phi}$ \cite{ijcai2018p97} combines local pixel values with the image-level mean value in one term, thus jointly capturing image-level statistics and local pixel matching information; and as standard, weighted F-measure $F_{\beta}^w$ and MAE (Mean Absolute Error).
\par The colors denoted in \textcolor{green}{green} and \textcolor{red}{red}
correspond to a \textit{relative} increase or decrease in the performance of
PDA and SDA, respectively, in comparison to the baseline methods, and this convention is followed in Table \ref{tab:table_bdd100k_qualitative_bdd_only}, \ref{tab:table_bdd100k_qualitative}, \ref{tab:table_cod_result} and \ref{tab:table_cod_camo_cham_result}. In the Quantitative section, we use the \textit{absolute} difference between baseline and PDA/SDA method by default unless specified explicitly.
\\
\paragraph{\noindent \textbf{Performance of EVP vs PDA and SDA - BDD100K}}
On the BDD100K dataset, SDA performs well as compared to EVP on $S_{\alpha}$, $E_{\phi}$ and $F_{\beta}^w$ by an average increase of 0.76$\%$ with the better MAE score as shown in the Table. \ref{tab:table_bdd100k_qualitative_bdd_only} . PDA performs well on $S_{\alpha}$, $E_{\phi}$ and $F_{\beta}^w$ by an average increase of 0.55$\%$, and with the best MAE score as compared to both EVP and SDA. Similarly, In the case of ACDC, Wild-Dash and Dark-Zurich datasets, SDA demonstrates significant performance gains as compared to EVP by 1.7$\%$, 1.2$\%$ and 2.67$\%$, with the best MAE scores, which demonstrates the generalization capability of SDA on the challenging adverse weather datasets. Also in terms of pure quantitative numbers shown in the Table. \ref{tab:table_bdd100k_qualitative}, the SDA model outperforms EVP and PDA on almost all the datasets.
\label{a}

\paragraph{\noindent \textbf{Performance of SAM-Adapter vs PDA (SAM Init.) and SDA (SAM Init.) - BDD100K}}
PDA is initialized by SAM ViT-B weights at the start of the training, and compared with SAM-Adapter with ViT-B architecture, trained on the BDD100K dataset. The Table. \ref{tab:table_bdd100k_qualitative_bdd_only}, PDA performs better on average at BDD100K and ACDC by 0.48$\%$ with the best MAE as compared to SAM-Adapter. In the Wild-Dash dataset, PDA has a performance gain of 0.94$\%$ with comparable MAE. In Table. \ref{tab:table_bdd100k_qualitative} Dark-Zurich there is an increase of 0.67$\%$ and 0.36$\%$ on $S_{\alpha}$ and $E_{\phi}$ with a decrease in $F_{\beta}^w$ score by 0.89$\%$, MAE score of SAM-Adapter is better than PDA in Dark-Zurich dataset. In most diverse settings PDA performance is better as compared to the SAM-Adapter. In the case of SDA with SAM-initialized weights, it performs 0.24$\%$ on the BDD100K dataset, and 0.54$\%$ on the ACDC dataset which is better than SAM-Adapter along with better MAE scores on both the datasets. On the Wild-Dash dataset SDA has a performance boost of 1.87 $\%$ with slightly worse MAE scores, while on the Dark-Zurich dataset, there is a performance decrease of almost 2.5 $\%$ along with the best MAE score as compared to SAM-Adapter. In terms of pure quantitative numbers in the Table. \ref{tab:table_bdd100k_qualitative_bdd_only} and \ref{tab:table_bdd100k_qualitative}, SDA is ahead of PDA and SAM-Adapter on most of the metrics on all the datasets.
\label{b}

\paragraph{\noindent \textbf{Performance of EVP vs PDA and SDA - COD and CAMO}}
In Table. \ref{tab:table_cod_result} and \ref{tab:table_cod_camo_cham_result}, the performance of PDA is comparable with EVP with a performance gain of 0.1$\%$ on the COD10K dataset and with a better MAE score as the best MAE score is achieved by SDA. While PDA gains 1.12$\%$ performance on the CAMO dataset with the best MAE score as compared to EVP. The observed performance enhancement in PDA can be attributed to its ability to capture high-frequency details which are learned by $\nabla HFC$ image processing block, such as edge information, which is crucial for achieving optimal performance in low-level semantic segmentation tasks, particularly on camouflage object segmentation datasets. PDA achieves this through the incorporation of a single $DiffAdaptor$ block at the onset of its architecture that helps it capture high-frequency details better than SDA. This sets it apart from older methods like EVP and SAM-Adapter, which use handcrafted $HFC$ blocks. \par In summary, our proposed SDA exhibits better average performance when compared to EVP, which is the current state-of-the-art (SOTA) method on BDD100K, ACDC, Wild-Dash, and Dark-Zurich datasets. In the case of SDA(SAM Init.) and PDA(SAM Init.), SDA(SAM Init.) outperforms the SAM-Adapter method. Additionally, PDA demonstrates strong performance in the CAMO and COD10K datasets compared to the EVP model. It is evident that our proposed methods, SDA and PDA, surpass the existing state-of-the-art (SOTA) methods, namely EVP and SAM-Adapter, in terms of generalization ability. Furthermore, there is a notable 51\%, 7\% and 55\%  \textit{relative} decrease in MAE when evaluating our methods on the Dark-Zurich, ACDC and BDD100k datasets, outperforming previous SOTA methods EVP and SAM-Adapter. The notable proficiency exhibited by SDA across datasets such as BDD100k, ACDC, Wild-Dash, and Dark-Zurich can be attributed to the paramount importance of integrating both local and global features throughout all hierarchical levels of the model, particularly within the domain of semantic segmentation tasks. Conversely, the performance of PDA on low-level semantic segmentation datasets such as CAMO and COD10K stems from the utilization of low-level features containing high-frequency feature information, which may significantly enhance performance outcomes.
\label{c}

\subsection{Segmentation Statistics}
In Fig.\ref{fig:stat_plot} we show through precision \& recall graph, how well our proposed SDA method performs against the EVP model on the contrary to evaluation metrics $S_{\alpha}$, $E_{\phi}$, $F_{\beta}^w$ and MAE.
\par In terms of precision, our method performs the best as compared to EVP which indicates that our method has fewer false positives in all the above test datasets. Recall of our system is better on BDD100K and Wild-Dash dataset while there is a slight drop in recall as the threshold increases in ACDC and Dark-Zurich.

\begin{table}[!htb]
\caption{\label{tab:table_cod_result} Comparison between EVP-Net and PDA/SDA on COD10K dataset with ViT-B as the backbone.}
\begin{adjustbox}{width=0.48\textwidth}
\begin{tabular}{|l|llll|}
\hline
\multicolumn{1}{|c|}{}                         & \multicolumn{4}{c|}{COD10K}                                                            \\
\multicolumn{1}{|c|}{\multirow{-2}{*}{Method}} & \multicolumn{1}{l|}{$S_{\alpha} \uparrow$}                             & \multicolumn{1}{l|}{$E_{\phi}\uparrow$}                             & \multicolumn{1}{l|}{$F_{\beta}^w \uparrow$}                            & $MAE\downarrow$                                                        \\ \hline
EVP-Net(VIT-B)                                 & \multicolumn{1}{l|}{.6180}                         & \multicolumn{1}{l|}{.6844}                         & \multicolumn{1}{l|}{\textbf{.3379}}      & .1089                     \\ \hline
  
PDA(ours)                      & \multicolumn{1}{l|}{\textbf{.6184} \textcolor{green}{(+0.06)}}    & \multicolumn{1}{l|}{\textbf{.6962 \textcolor{green}{(+1.72)}}}    & \multicolumn{1}{l|}{.3286 \textcolor{red}{(-2.83)}}    &  .1005 \textcolor{green}{(+8.3)}                                \\ \hline

SDA(ours)                      & \multicolumn{1}{l|}{.618}    & \multicolumn{1}{l|}{.695 \textcolor{green}{(+1.55)}}    & \multicolumn{1}{l|}{.3305 \textcolor{red}{(-2.24)}}    & \textbf{.1003} \textcolor{green}{(+8.5)}   \\ \hline
\end{tabular}
\end{adjustbox}
\end{table}


\begin{table}[!htb]
\caption{\label{tab:table_cod_camo_cham_result} Comparison between EVP-Net and PDA/SDA on CAMO dataset with ViT-B as the backbone.}
\begin{adjustbox}{width=0.48\textwidth}
\begin{tabular}{|l|llll|}
\hline
\multicolumn{1}{|c|}{}                         & \multicolumn{4}{c|}{CAMO}                                                            \\
\multicolumn{1}{|c|}{\multirow{-2}{*}{Method}} & \multicolumn{1}{l|}{$S_{\alpha} \uparrow$}                             & \multicolumn{1}{l|}{$E_{\phi}\uparrow$}                             & \multicolumn{1}{l|}{$F_{\beta}^w \uparrow$}                            & $MAE\downarrow$                                                        \\ \hline
EVP-Net(VIT-B)                                 & \multicolumn{1}{l|}{ .4994} & \multicolumn{1}{l|}{ .6151} & \multicolumn{1}{l|}{ .2496} &  .2039                      \\ \hline
  
PDA(ours)                      & \multicolumn{1}{l|}{\textbf{.5173} \textcolor{green}{(+3.58)}}    & \multicolumn{1}{l|}{\textbf{.6176 \textcolor{green}{(+0.41)}}}    & \multicolumn{1}{l|}{\textbf{.2628} \textcolor{green}{(+5.29)}}    & \textbf{.1937}     \textcolor{green}{(+5.3)}                           \\ \hline

SDA(ours)                      & \multicolumn{1}{l|}{.5002 \textcolor{green}{(+0.16)}}    & \multicolumn{1}{l|}{.6156 \textcolor{green}{(+0.08)}}    & \multicolumn{1}{l|}{.243 \textcolor{red}{(-8.15)}}    & .1965 \textcolor{green}{(+3.8)} \\ \hline
\end{tabular}
\end{adjustbox}
\end{table}

\begin{table}[!htb]
\caption{\label{tab:table_dvp_abalation}Abalation for PDA design choice.}
\begin{adjustbox}{width=0.48\textwidth}
\begin{tabular}{|l|llll|}

\hline
\multicolumn{1}{|c|}{}                         & \multicolumn{4}{c|}{BDD100K}                                                                                                                                                                   \\
\multicolumn{1}{|c|}{\multirow{-2}{*}{Method}} & \multicolumn{1}{l|}{$S_{\alpha} \uparrow$}                             & \multicolumn{1}{l|}{$E_{\phi}\uparrow$}                             & \multicolumn{1}{l|}{$F_{\beta}^w \uparrow$}                            & $MAE\downarrow$                             \\ \hline
Vanilla ViT-B & \multicolumn{1}{l|}{.8454}    & \multicolumn{1}{l|}{.9217}    & \multicolumn{1}{l|}{.7542}    & .0316     \\ \hline
EVP                                 & \multicolumn{1}{l|}{.8466} & \multicolumn{1}{l|}{.9225} & \multicolumn{1}{l|}{.7521} & .03033     \\ \hline

PDA (w/o $f_{emb}$)                                 & \multicolumn{1}{l|}{.8479} & \multicolumn{1}{l|}{.9229} & \multicolumn{1}{l|}{.7534} & .03065    \\ \hline

PDA (w/o $DiffIP$)                                & \multicolumn{1}{l|}{.846} & \multicolumn{1}{l|}{.9222} & \multicolumn{1}{l|}{.7515} & .03082    \\ \hline

PDA (w/o $DiffIP$ \& w/o $f_{emb}$) & \multicolumn{1}{l|}{.8449}    & \multicolumn{1}{l|}{.9213}    & \multicolumn{1}{l|}{.7492}    & .03109     \\ \hline

 PDA (ours)                      & \multicolumn{1}{l|}{\textbf{.8513}}    & \multicolumn{1}{l|}{\textbf{.9270}}    & \multicolumn{1}{l|}{\textbf{.759}}    & \textbf{.02940}     \\ \hline
  
\end{tabular}

\end{adjustbox}

\end{table}

\begin{table}[!htb]
\caption{\label{tab:hfc_abalation} Comparision between EVP with only HFC, $\nabla$HFC with $f_{emb}$.}
\begin{adjustbox}{width=0.48\textwidth}
\begin{tabular}{|l|llll|}
\hline
\multicolumn{1}{|c|}{}                         & \multicolumn{4}{c|}{BDD100K}                                                                                                                                                                   \\
\multicolumn{1}{|c|}{\multirow{-2}{*}{Method}} & \multicolumn{1}{l|}{$S_{\alpha} \uparrow$}                             & \multicolumn{1}{l|}{$E_{\phi}\uparrow$}                             & \multicolumn{1}{l|}{$F_{\beta}^w \uparrow$}                            & $MAE\downarrow$                             \\ \hline
EVP                                 & \multicolumn{1}{l|}{.8466} & \multicolumn{1}{l|}{.9225} & \multicolumn{1}{l|}{.7521} & .03033     \\ \hline
$\nabla$HFC                      & \multicolumn{1}{l|}{.847}    & \multicolumn{1}{l|}{.9238}    & \multicolumn{1}{l|}{.7526}    & .03073     \\ \hline
$\nabla$HFC + $f_{emb}$                      & \multicolumn{1}{l|}{\textbf{.8512}}    & \multicolumn{1}{l|}{\textbf{.9245}}    & \multicolumn{1}{l|}{\textbf{.7584}}    & \textbf{.02965}     \\ \hline
  
\end{tabular}
\end{adjustbox}

\end{table}

\subsection{Ablation Studies}
\label{sec:abalation}
We conducted an ablation study on our PDA model, utilizing the BDD100K dataset for validation purposes. This rigorous examination of our design decisions is undertaken due to the deliberate inclusion of only one $DiffVP$ block within the model, a simplification intended to facilitate a more comprehensive understanding of its performance characteristics and details of which can be referred to in Table \ref{tab:table_dvp_abalation}.

\paragraph{\noindent PDA v/s PDA(w/o $DiffIP$ \& w/o $f_{emb}$) }
PDA is better than PDA(w/o $DiffIP$ \& w/o $f_{emb}$) and shallow vision encoder by 0.73$\%$ on an average on $S_{\alpha}$, $E_{\phi}$ and $F_{\beta}^w$ with best MAE scores. In this setup, the image is directly passed through the $VPTune$ block without any pre-processing on it.

\paragraph{\noindent PDA v/s PDA(w/o $DiffIP$) }
PDA is better than PDA(w/o $DiffIP$) by 0.6$\%$ on an average on BDD100K dataset with the best MAE scores which tells that (0.73$\%$ - 0.6$\%$) = 0.1$\%$ gap is closed by $f_{emb}$ and is evident from the above abalation.

\paragraph{\noindent PDA v/s PDA(w/o $f_{emb}$) }
As shown in Table \ref{tab:table_dvp_abalation}, PDA performance is better than in PDA(w/o $f_{emb}$) by 0.44 $\%$ and has better MAE scores. It is evident that 
the performance gain by having $DiffIP$ is about (0.73$\%$ - 0.44$\%$) = 0.29$\%$. 
\\
Therefore, PDA with both $DiffIP$ and $f_{emb}$ is the best setting for training rather than PDA(w/o $DiffIP$) and PDA(w/o $f_{emb}$). And by jointly learning $DiffIP$ and $f_{emb}$ there is a performance gain of additional (0.73$\%$ - (0.29$\%$+0.1$\%$)) = 0.34$\%$ and while the individual contribution of $DiffIP$ and $f_{emb}$ is 0.29$\%$ and 0.1$\%$.
\\
\noindent We also conduct further ablation on the effectiveness of our $\nabla HFC$ Image processing block as compared to EVP HFC block as shown in Table \ref{tab:hfc_abalation}.

\paragraph{\noindent $HFC$ vs $\nabla HFC$}
PDA with only $\nabla HFC$ the performance gain is 0.07 $\%$ with comparable MAE scores. So $\nabla HFC$ will give a slight performance gain in the PDA model as compared to the fixed $HFC$ block in the case of EVP.

\paragraph{\noindent $HFC$ vs $\nabla HFC$ + $f_{emb}$}
PDA with $\nabla HFC$ and $f_{emb}$ jointly increase the performance by 0.43$\%$ as compared to EVP with $HFC$ block.

To summarize, using PDA with $f_{emb}$ and $DiffIP$ containing $\nabla HFC$ gives the performance benefit and is better than the state-of-the-art EVP and SAM-Adapter models.

\section{CONCLUSION}
\label{sec:conclusion}

In this paper, we have introduced DiffPrompter, a novel differentiable visual and latent prompting mechanism for adaptors aimed at addressing semantic segmentation tasks. Our study concludes that the integration of the $\nabla$HFC image processing block alongside our series (SDA) architectures exhibits exceptional performance in vehicle segmentation across datasets featuring adverse weather conditions. Conversely, the parallel (PDA) architecture demonstrates proficiency in low-level semantic segmentation tasks. Furthermore, our ablation study validates the efficacy of concurrently learning image and latent embeddings, directly correlating with enhanced performance in segmentation tasks and achieving state-of-the-art results in out-of-distribution generalization.
\par 
In the future, we aim to delve deeper into the exploration of visual prompts and latent prompts within the realm of object detection tasks, extending our investigations to encompass the domain of 3D scene understanding. Analogous to their role in language models, language prompts offer a discrete structure that has proven instrumental in achieving optimal results in language model generation. Consequently, directing attention towards the processing of visual prompts, with a specific emphasis on capturing unstructured elements inherent to the scene, emerges as a pivotal objective. By prioritizing this endeavour, we anticipate advancing towards the development of a more robust visual foundation model, thereby enhancing our comprehension and interpretation of complex visual environments which will help solve challenging robotic tasks directly.

\bibliography{root}
\bibliographystyle{IEEEtran}
\end{document}